\documentclass[conference]{IEEEtran}
\IEEEoverridecommandlockouts
\usepackage{cite}
\usepackage{amsmath,amssymb,amsfonts}
\usepackage{algorithmic}
\usepackage{graphicx}
\usepackage{textcomp}
\usepackage{xcolor}
\usepackage{tabularx}
\usepackage{hyperref}
\usepackage{tikz}
\usetikzlibrary{fit, shapes, positioning, patterns}
\usepackage{subcaption} 
\def\BibTeX{{\rm B\kern-.05em{\sc i\kern-.025em b}\kern-.08em
    T\kern-.1667em\lower.7ex\hbox{E}\kern-.125emX}}
\begin{document}

\newcommand\copyrighttext{%
	\footnotesize \copyright{ }2023 IEEE. Personal use of this material is permitted. Permission from IEEE must be obtained for all other uses, in any current or future media, including reprinting/republishing this material for advertising or promotional purposes, creating new collective works, for resale or redistribution to servers or lists, or reuse of any copyrighted component of this work in other works.}
\newcommand\copyrightnotice{%
	\begin{tikzpicture}[remember picture,overlay]
	\node[anchor=south,yshift=10pt,xshift=7pt] at (current page.south) {\parbox{\dimexpr\textwidth-\fboxsep-\fboxrule\relax}{\copyrighttext}};
	\end{tikzpicture}%
}

\title{Framework for Quality Evaluation of Smart Roadside Infrastructure Sensors for Automated Driving Applications\\
\thanks{This research is accomplished within the project ”AUTOtech.\textit{agil}” (FKZ 01IS22088A). We acknowledge the financial support for the project by the Federal Ministry of Education and Research of Germany (BMBF).

\IEEEauthorrefmark{1}C. Liu and C. Wei assert joint second authorship.}
}

\makeatletter
\newcommand{\linebreakand}{%
  \end{@IEEEauthorhalign}
  \hfill\mbox{}\par
  \mbox{}\hfill\begin{@IEEEauthorhalign}
}
\makeatother

\author{\IEEEauthorblockN{Laurent Kloeker}
\IEEEauthorblockA{\textit{Institute for Automotive Engineering} \\
\textit{RWTH Aachen University}\\
Aachen, Germany\\
laurent.kloeker@ika.rwth-aachen.de}
\and
\IEEEauthorblockN{Chenghua Liu\IEEEauthorrefmark{1}}
\IEEEauthorblockA{\textit{Institute for Automotive Engineering} \\
\textit{RWTH Aachen University}\\
Aachen, Germany \\
chenghua.liu@rwth-aachen.de}
\and
\IEEEauthorblockN{Chao Wei\IEEEauthorrefmark{1}}
\IEEEauthorblockA{\textit{Institute for Automotive Engineering} \\
\textit{RWTH Aachen University}\\
Aachen, Germany \\
chao.wei@rwth-aachen.de}
\linebreakand
\IEEEauthorblockN{Lutz Eckstein}
\IEEEauthorblockA{\textit{Institute for Automotive Engineering} \\
\textit{RWTH Aachen University}\\
Aachen, Germany \\
lutz.eckstein@ika.rwth-aachen.de}
}
\maketitle

\copyrightnotice

\begin{abstract}
The use of smart roadside infrastructure sensors is highly relevant for future applications of connected and automated vehicles. External sensor technology in the form of intelligent transportation system stations (ITS-Ss) can provide safety-critical real-time information about road users in the form of a digital twin. The choice of sensor setups has a major influence on the downstream function as well as the data quality. To date, there is insufficient research on which sensor setups result in which levels of ITS-S data quality. We present a novel approach to perform detailed quality assessment for smart roadside infrastructure sensors. Our framework is multimodal across different sensor types and is evaluated on the DAIR-V2X dataset. We analyze the composition of different lidar and camera sensors and assess them in terms of accuracy, latency, and reliability. The evaluations show that the framework can be used reliably for several future ITS-S applications. 
\end{abstract}

\begin{IEEEkeywords}
Smart roadside infrastructure sensors, framework, quality assessment, digital twin, automated driving
\end{IEEEkeywords}

\section{INTRODUCTION}
The implementation of smart roadside infrastructure sensors is crucial for the successful integration and utilization of connected and automated vehicles (CAVs) in the future. In urban areas in particular, high traffic densities, buildings or vegetation pose major technical challenges for CAVs, as they are unable to adequately detect and interpret the entire environment under such circumstances~\cite{Tas2018}. In this context, external sensor technology in the form of intelligent infrastructure, or intelligent transportation system stations (ITS-Ss), enable the collection of safety-critical information about the current state of other road users in the form of a digital twin~\cite{Kloeker2021}. Here, a digital twin means the transformation of a physical system in a digital environment. This state information can include, for example, the position, dimensions, and velocity of each road user and be transmitted to CAVs via infrastructure-to-vehicle (I2V) communication. The added value thus gained is reflected in an enhanced and more robust environmental perception for CAVs which results in a more reliable trajectory planning~\cite{Pechinger2021}. Recorded traffic data can also be transferred to an offline simulation environment apart from the online operation and used for the development and validation of automated driving functions~\cite{Glasmacher2023}. In addition to specific applications in the field of automated driving, ITS-Ss can also be additionally located in the thematic field of smart cities. As an example, sensor technology can be used to monitor and improve traffic flow in real-time, e.g. by adjusting traffic lights. Another application is monitoring parking lots to detect the availability of free parking spaces and help drivers find them.

The downstream function of an ITS-S application and the associated data quality are highly dependent on the applied sensor technology. Different downstream functions have different requirements for the accuracy, latency, and reliability of sensor data processing. As previously described, use cases range from coarse traffic counting to high-accuracy trajectory extraction. To date, however, there is insufficient evidence on what sensor technology results in what levels of data quality. Thus, it is not directly comprehensible for future ITS-S operators which sensor qualities have to be procured at which costs.

In this paper, we present a novel approach to perform a detailed quality assessment for smart roadside infrastructure sensors. Our framework is multimodal across different sensor types and is evaluated using the DAIR-V2X dataset~\cite{yu2022dairv2x} as an example of lidar and camera sensing. We evaluate the compilation of single and multi-sensor setups and types per virtual recording location at different sensor resolutions and with different computational hardware available. This is followed by a holistic evaluation with respect to the accuracy, latency, and reliability of the respective setup. All evaluation steps are substantially related to the automated detection and tracking of road users.

\section{RELATED WORK}
To the best of our knowledge, no current systematic review on the quality assessment of smart roadside infrastructure sensors for automated driving applications exists. Bai et al.~\cite{bai2022infrastructure} focus on infrastructure-based object detection and tracking analysis but do not explicitly analyze different sensor setups and sensor qualities. However, the following subsections illustrate the considerable need for such investigations on the basis of current research activities.

\subsection{Current Activities on ITS-Ss}
ITS-Ss have already been operated for several years for various purposes in real traffic. In~\cite{Cress2021}, a list of the most prominent ITS-S activities in the field of connected and automated mobility on a global scale is given. It is noticeable that especially recent ITS-S activities have been set up in the form of digital test fields and rely on the combined use of camera, radar and lidar sensor technology to detect road users. The main purpose of these digital test fields is to verify and validate automated and connected driving functions as well as the infrastructure-enabled operation of CAVs~\cite{Zofka2022}. Examples of significant activities in Germany are the ACCorD~\cite{Kloeker2021} and Providentia~\cite{Kraemmer2019} digital test fields. While the ACCorD test field uses a mix of over 250 camera and lidar sensors, the Providentia test field uses a mix of 75 camera, radar and lidar sensors. Both projects aim at transferring the traffic events of the measurement cross-sections into a digital twin with highest accuracy and reliability at minimum latency. For this purpose, sensors with the highest resolution -- available at the time the digital test fields were created -- were used.

\subsection{Smart Roadside Infrastructure Sensor Datasets} \label{IIB}
Sensor datasets from the vehicle perspective are available in large numbers in the research area of CAVs and are already adequately known, e.g.~\cite{geiger2013vision, sun2020scalability}. Sensor datasets from an infrastructure perspective, in contrast, have been almost non-existent to date. However, the aforementioned increasing activities in the field of ITS-S prove the acute need for such datasets in order to conduct more advanced research in this area in the future. Sun et al. provide an overview of the only five infrastructure datasets currently available~\cite{Sun2022}: DAIR-V2X~\cite{yu2022dairv2x}, BAAI-VANJEE~\cite{yongqiang2021baai}, IPS300+~\cite{wang2022ips300+}, A9-Dataset~\cite{cress2022a9} and LUMPI~\cite{busch2022lumpi}. All datasets consist of a combination of at least one camera sensor and one lidar sensor. Radar sensors are not represented. The characteristics of the sensors used varies widely. Except for the LUMPI dataset, there is always at least one pair consisting of a camera and lidar sensor at approximately the same position with an equally similar field of view (FOV). This is a relevant property for the intermodal evaluation of sensor types and qualities.

\section{METHOD}\label{III}
\subsection{Requirements for the Assessment Framework} \label{IIIA}
The framework for evaluating smart roadside infrastructure sensors must meet several requirements in order to be able to make multimodal and holistic statements. For this reason, the following seven requirements are placed on the underlying methodology:
The framework must ...
\begin{enumerate}
\item ... be independent of the sensor technology used and be able to handle it on a modular basis. 
\item ... be able to make statements about different sensor concepts. 
\item ... be independent of the used computing hardware and consider its influence on the results. 
\item ... be usable by different actors with different requirement profiles and downstream functions. 
\item ... be able to make generalistic statements independent of the downstream function. 
\item ... be tested and evaluated using a data-driven approach. 
\end{enumerate}

\subsection{Framework Architecture}
The task of the framework architecture is to fulfill all requirements defined in subsection~\ref{IIIA} and to generate reproducible results. Figure~\ref{fig:framework-architecture} provides an overview of the underlying architecture.
\begin{figure*}[t]
  \includegraphics[width=\textwidth]{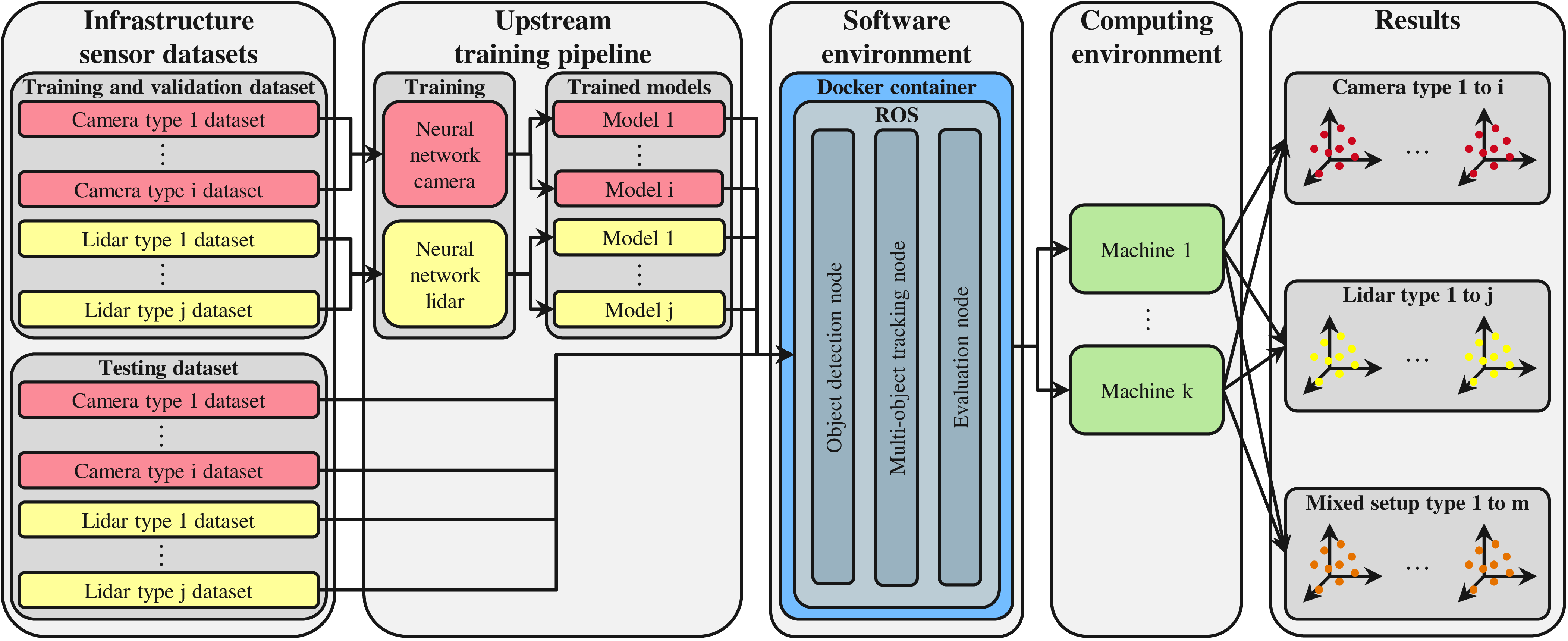}
  \caption{Framework architecture for quality evaluation of smart roadside infrastructure sensors.}
  \label{fig:framework-architecture}
\end{figure*}

The input consists of any infrastructure sensor dataset of any sensor type with any quality and resolution. 
Based on the requirements for the sensor data sets used, which are defined in subsection~\ref{IIIC}, figure~\ref{fig:framework-architecture} already focuses on the use of a camera and a lidar sensor data set. For each sensor type and resolution, a representative training and validation dataset as well as a test dataset must be provided. Then, matching neural network architectures are trained with the corresponding training datasets and one trained model per sensor is returned. It is important to ensure the interchangeability of the different trained models in order to seamlessly integrate new, more powerful neural network architectures into the framework. For this purpose, we make use of a training platform that has not yet been published and that enables exactly this. In our case, we use the PBOD architecture~\cite{wang2020pillar} for lidar-related tasks and the SMOKE architecture~\cite{liu2020smoke} for camera-related tasks. Both network architectures generate three-dimensional bounding boxes, a required condition for subsequent cross-sensor quality assessment.

The software environment consists of a \emph{Docker} container with an embedded \emph{Robot Operating System (ROS)} environment. First, multi-object detection is performed with each related combination of sensor-specific dataset and trained model. The datasets are transferred to ROS-bags beforehand and played at a low playback rate to avoid data congestion and loss and thus falsified results. With eight different camera resolutions \(C_i\) and six different lidar resolutions \(L_j\), we end up with 14 such jobs. Subsequently, multi-object tracking is performed for each single sensor setup as well as for each cross-sensor type combination of all resolution levels. With
\begin{equation}
    {n_{sensor,combinations}} = i+j+i \cdot j
\end{equation}
this results in 62 sensor combinations in our case.

The underlying concept of using a Docker container is the possibility of including different computational environments for hardware related evaluations. In real use cases, sensor data is evaluated by \mbox{ITS-Ss} on local or cloud-hosted computing hardware. In order to also be able to make statements about the influence of different computational hardware qualities on the overall result, the Docker container is run on different machines. In our case, four different workstations and servers are used, whose specifications are given in table~\ref{tab:machine_specs}.

\begin{table}[t]
\caption{GPU and CPU specifications of the computing hardware used.}
\label{tab:machine_specs}
\begin{center}
\begin{tabularx}{86 mm}{ |l||X||X| }
  \hline
  \textbf{Machine no.} & \textbf{GPU} & \textbf{CPU}\\
  \hline 
  1 & Dual Nvidia A100 40~GB & Dual AMD Epyc 7742\\
  \hline
  2 & Nvidia A100 40~GB & Dual AMD Epyc 7742\\
  \hline
  3 & Nvidia RTX 3090 24~GB & AMD Ryzen Threadripper 3960X\\
  \hline
  4 & Nvidia Quadro RTX 6000 24~GB & AMD Ryzen 7 3700X\\
  \hline
\end{tabularx}
\end{center}
\vspace{-20pt}
\end{table}

Evaluating the 62 sensor combinations on four different machines, we finally obtain \(n=248\) total combinations. For each total combination, a holistic evaluation is then performed, resulting in the quality assessment results.

\subsection{Used Road Side Infrastructure Sensor Dataset} \label{IIIC}
As presented in section~\ref{IIB}, five publicly available infrastructure sensor datasets exist to date. Based on the previously formulated requirements, additional specific requirements for the choice of the dataset are derived. The dataset to be used must consist of at least one camera sensor and one lidar sensor. Sensors of different types must record from nearly identical positions and FOVs to ensure interchangeability. In addition, all object annotations of the dataset must have information in three-dimensional space to ensure interchangeability at the object level as well. The coordinate transformation between the different sensor coordinate systems must be known. Since the framework aims to evaluate not only different sensor types but also different sensor qualities, multiple resolution levels per sensor must be considered. However, no dataset contains sensor information of different resolutions at the same position and FOV. For this reason, we will gradually scale down the existing sensor resolutions to simulate different quality levels.

LUMPI and A9-dataset already do not meet the requirements regarding equal sensor position and FOV. BAAI-VANJEE is also not suitable since the lidar sensors used only have a vertical resolution of 32 layers and are thus only suitable for a few further downscaling steps. The lidar sensors of IPS300+ have 80 vertical layers and the most annotated frames, but the quality of the annotations and the coordinate transformations between the sensors are of too low quality to be suitable for meaningful and reproducible analyses. Accordingly, only the DAIR-V2X dataset remains. It consists of 10,084 jointly annotated frames recorded with a 300 vertical layers lidar sensor and a 1920x1080 pixel resolution camera at an urban intersection in Beijing, China. The specifications of the sensors used can be found in table~\ref{tab:sensor_specs}.

\begin{table}[t]
\caption{Specifications of the infrastructure sensors used in the DAIR-V2X dataset.}
\label{tab:sensor_specs}
\begin{center}
\begin{tabularx}{86 mm}{ |p{18mm}||X| }
  \hline
  \textbf{Sensor} & \textbf{Details}\\
  \hline 
  Lidar: InnoVusion Jaguar Prime & 10~Hz sample rate, 300 vertical beams, 100$^\circ$ hor. FOV, 40$^\circ$ vert. FOV, 0.09-0.33$^\circ$ hor. angular resolution, 0.13$^\circ$ vert. angular resolution, ±3~cm distance accuracy\\
  \hline
  Camera: HIKVISION iDS-TCE900QX-B & Sony IMX267LLR CMOS sensor with RGB filter, 25~Hz sample rate, 1920x1080~px resolution, 48.1$^\circ$ hor. FOV, 27.7$^\circ$ vert. FOV, 15.9~mm focal length, global shutter, JPEG compressed\\
  \hline
\end{tabularx}
\end{center}
\vspace{-20pt}
\end{table}

Based on the available raw sensor data, resolution levels are defined to simulate possible infrastructure sensors that may be used in current or future applications. Camera: \(C_i=\{2160,1080,720,540,360,270,180,135\}\) vertical pixels, with a respective aspect ratio of 16:9. Lidar: \(L_j=\{256,128,64,32,16,8\}\) vertical planes. Since no information about the associated vertical layer per point is stored in the lidar point clouds, we apply our own algorithm to identify the layer membership of individual points. In the first downsampling step, the topmost 44 layers are first removed to reach the highest resolution of 256 layers for our investigations. Subsequently, every second layer is removed, while steadily maintaining the lowest layer in each downsampling step. This ensures that the remaining layers in each downsampling step represent the relevant near range instead of the less relevant far range. The camera images are at 1080 vertical pixels, so we need to perform both upsampling and downsampling steps. Upsampling to 2160 vertical pixels is done using nearest neighbor value interpolation~\cite{Rukundo2012} and downsampling to all remaining resolutions is done using a Lanczos filter~\cite{madhukar2013lanczos}.

In the next step, both the horizontal and vertical FOV of the lidar sensor must be aligned with those of the camera sensor in order to analyze only identical objects in both sensor data per frame. The cropping of the lidar point clouds takes place based on the determination of feature points. The DAIR-V2X dataset consists of a training dataset and a test dataset. The test dataset differs from the training dataset in that it contains 85 different sequences whose frames are chronologically related. The frames of the training dataset are not chronologically connected and are therefore not suitable for possible object tracking analyses. A major disadvantage of the test dataset, however, is the annotations provided, which were not generated manually but by a detector and are of very poor quality. Thus, in order to perform reliable evaluations regarding object detection and tracking, the frames of the test dataset have to be corrected manually. Due to the high time effort involved, we decided to use three representative sequences. These three sequences contain all object classes, different weather conditions and balanced motion flows of the objects in all directions. After splitting the training dataset into a training and validation dataset, we obtain 5988 frames for training, 2511 frames for validation, and 472 frames for testing. Moreover, we reduce the seven original classes of the DAIR-V2X dataset to the four most common ones: pedestrian, bike (mix of bicycle and scooter), car (mix of car and van), and truck (mix of truck and bus). For the evaluation of detected objects, the temporal offset between lidar and camera frames must always be considered in the next steps, due to the non-synchronized original acquisition rates of 25~Hz (camera) and 10~Hz (lidar). Although the camera frames are present in the dataset at a virtual frequency of 10~Hz, the associated time stamps correspond to the time stamps of the lidar frames.

\subsection{Metrics for Quality Assessment}
Looking at different use cases of ITS-Ss, we have identified three key performance indices (KPIs): \emph{Accuracy}, \emph{Latency}, and \emph{Reliability}. Accuracy determines the quality of extracted objects and trajectories from raw sensor data. Latency is an important indicator to check whether the setup is suitable for real-time applications or not. Reliability, in turn, indicates how consistently the accuracy and latency values can be maintained. Not every application allows large fluctuations around an intended average value. The aim is to represent these three values on a normalized scale between 0 and 1, bundled in a quality vector \(Q_n\).

The accuracy of a system depends on the sensor accuracy \(A_s\), on the localization accuracy of the sensor \(A_l\), on the object detection accuracy \(A_d\) and on the tracking accuracy \(A_t\). To determine \(A_{s_{i,j}}\), we first define a sensor-independent maximum detection distance of \(x_{detection} = 150 m\). In addition, we need to identify the type-dependent sensor error \(e_{s_{i,j}}\), which depends on the ground sampling distance (GSD) in image width \(w\) and image height \(h\) for camera sensors and on the beam uncertainty for lidar sensors. Here we invoke the formulaic relationships from~\cite{kloeker2022generic}.
\begin{equation}
    A_{s_{i,j}} = 1-\frac{e_{s_{i,j}}}{x_{detection}}
\end{equation}
with
\begin{equation}
    e_{s_{i}} = max(GSD_{w,h})
\end{equation}
\begin{equation}
    e_{s_{j}} = E_{LiDAR}(x_{detection})
\end{equation}
\(A_{l_{i,j}}\) depends on the initial and continuous registration of the local sensor coordinate system into a global reference coordinate system. It consists of a translational and a rotational part. In general, these values are constant over a measurement period. In~\cite{lv2021cfnet} and~\cite{kloeker2020real}, values for camera and lidar sensor systems based on real measurements have already been determined. Here we use the values for measurement cross sections that are closest to our use case: \(e_{trans_{i}}=0.519cm\), \(e_{rot_{i}}=0.09^{\circ}\), \(e_{trans_{j}}=4cm\) and \(e_{rot_{j}}=0.03^{\circ}\).
\begin{equation}
    A_{l_{i,j}} = 1-\frac{e_{l_{i,j}}}{x_{detection}}
\end{equation}
with
\begin{equation}
    e_{l_{i,j}} = \sqrt{e_{trans_{i,j}}^2+(x_{detection}*e_{rot_{i,j}})^2}
\end{equation}
To determine detection accuracy, we use the widely used mean average precision (mAP) metric at an intersection over union (IoU) threshold of 0.5.
\begin{equation}
    A_{d_{i,j}} = mAP@[0.5]
\end{equation}
The composite accuracy \(A_{s,l,d}\) of a sensor is now given by
\begin{equation}
    (A_{s,l,d})_{i,j} = A_{s_{i,j}} * A_{l_{i,j}} * A_{d_{i,j}}
\end{equation}
For combined sensor setups \(m\) applies:
\begin{equation}
\begin{split}
    (A_{s,l,d})_{m} = \frac{(A_{s,l,d})_{i}}{(A_{s,l,d})_{i}+(A_{s,l,d})_{j}}*(A_{s,l,d})_{i}+\\
    \frac{(A_{s,l,d})_{j}}{(A_{s,l,d})_{i}+(A_{s,l,d})_{j}}*(A_{s,l,d})_{j}
\end{split}
\end{equation}
To evaluate \(A_t\), we use the Higher Order Tracking Accuracy (HOTA) of Luiten et al~\cite{luiten2021hota}. In doing so, we extend the evaluation of the 2D IoU in the HOTA algorithm for our use case by a 3D IoU.
\begin{equation}
    A_{t_{i,j}} = HOTA_{i,j}
\end{equation}
\begin{equation}
    A_{t_{m}} = \frac{A_{t_{i}}+A_{t_{j}}}{2}
\end{equation}
The final and already normalized accuracy thus results in
\begin{equation}
    Accuracy_{norm_{n}} = \sqrt[4]{(A_{s,l,d})_{n}*A_{t_{n}}} \in [0;1]
\label{eq:A}
\end{equation}

The latency of a system is the elapsed time from the recording of a sensor frame to the completed execution of object detection and tracking of all objects located in the frame. It is significantly dependent on the computing hardware and sensor resolution used. While object detection runs mainly on the GPU, object tracking is handled by the CPU.
\begin{equation}
\begin{split}
    Latency_{n} = t_{sensor exposure_{n}} + t_{readout_{n}} + \\
    t_{local network transmission_{n}} + t_{I/O operation_{n}} + \\
    t_{detection_{n}} + t_{tracking_{n}}
\end{split}
\end{equation}
Due to the not concretely feasible differentiation of the first four summands, we summarize them simplified as
\begin{equation}
\begin{split}
    t_{sensor readout_{n}} = t_{sensor exposure_{n}} + t_{readout_{n}} + \\
    t_{local network transmission_{n}} + t_{I/O operation_{n}} \\
\end{split}
\end{equation}
Normalization of the total latency is performed using a minimum optimum value \(t_{min}=0 ms\) and a maximum acceptable upper limit for real-time ITS-S applications of \(t_{max}=1000 ms\). :
\begin{equation}
    Latency_{norm_{n}} = 1-\frac{Latency_{n}-t_{min}}{t_{max}-t_{min}} \in [0;1]
\label{eq:L}
\end{equation}
with
\begin{equation}
    Latency_n= 
    \begin{cases}
        Latency_{n}, & \text{for } 0 < Latency_{n} < t_{max}\\
        t_{max}, & \text{for } Latency_{n} \geq t_{max}\\ 
    \end{cases}
\end{equation}

We represent the reliability of a system by the variances of dataset characteristics, accuracies, and latencies.
\begin{equation}
    Reliability_{n} = Var(R_{1_n} + R_{2_n} + R_{3_n} + R_{4_n} + R_{5_n})
\end{equation}
with
\begin{equation}
    R_{1_n} = (\frac{n_{objects}}{frame})_{n}
\end{equation}
\begin{equation}
    R_{2_n} = (\frac{A_{d}}{frame})_{n}
\end{equation}
\begin{equation}
    R_{3_n} = (\frac{t_{tracking}}{frame})_{n}
\end{equation}
\begin{equation}
    R_{4_n} = (\frac{t_{detection}}{frame})_{n}
\end{equation}
\begin{equation}
    R_{5_n} = ({A_{t}})_{n}
\end{equation}
Investigations have shown that correlations exist exclusively between \(R_{1}\) and \(R_{2}\), and \(R_{1}\) and \(R_{3}\). Due to the fact that we only get one output per sequence with the HOTA algorithm, no variance for \(R_{5}\) can be derived. Thus, the non-normalized reliability results as follows
\begin{equation}
\begin{split}   
    Reliability_{n} = Var(R_{1_n}) + Var(R_{2_n}) + Var(R_{3_n}) +\\
    Var(R_{4_n}) + 2Cov(R_{1_n},R_{2_n}) + 2Cov(R_{1_n},R_{3_n})
\end{split}
\end{equation}
The normalization is obtained by relating the current reliability value to the maximum and minimum reliability value of all \(n=248\) total combinations.
\begin{equation}
\begin{split}
    Reliability_{norm_{n}} =\\
    1-\frac{Reliability_{n}-min(Reliability_n)}{max(Reliability_n)-min(Reliability_n)} \in [0;1]
\end{split}
\label{eq:R}
\end{equation}
All now normalized values (\ref{eq:A}), (\ref{eq:L}) and (\ref{eq:R}) of a total combination can then be transformed into a final quality vector.
\begin{equation}
    Q_n = \begin{pmatrix}Accuracy_{norm_{n}}\\Latency_{norm_{n}}\\Reliability_{norm_{n}}\end{pmatrix}, n \in \{i,j,m\}
\end{equation}

\section{RESULTS}
After determining the metrics for generating the quality vector, all \(n=248\) total combinations are processed in the framework. In the following subsections, accuracy, latency, reliability and the final quality vector are analyzed qualitatively and quantitatively. The goal is to validate the framework derived in section~\ref{III} on the present DAIR-V2X dataset.
\begin{table}[t]
\caption{Listing of accuracy metrics for selected sensor setups. The calculation has taken place on machine no. 1.}
\label{tab:mAP}
\begin{center}
\begin{tabularx}{86 mm}{ |l||X|X|X|X| }
  \hline
  \textbf{Setup} & \textbf{mAP} & \textbf{\(A_{s,l,d}\)} & \textbf{HOTA} & \textbf{\(A_{norm}\)}\\
  \hline 
  C2160 & 0.0819 & 0.0817 & 0.1032 & 0.3030\\
  \hline
  C540 & \textbf{0.0859} & \textbf{0.0857} & \textbf{0.1290} & \textbf{0.3242}\\
  \hline
  C135 & 0.0046 & 0.0046 & 0.0597 & 0.1287\\
  \hline
  \hline
  L256 & \textbf{0.6243} & \textbf{0.6237} & \textbf{0.4270} & \textbf{0.7184}\\
  \hline
  L32 & 0.4454 & 0.4450 & 0.2904 & 0.5996\\
  \hline
  L8 & 0.2464 & 0.2462 & 0.1511 & 0.4392\\
  \hline
  \hline
  C2160 \& L256 & \textbf{0.5964} & \textbf{0.5959} & \textbf{0.2128} & \textbf{0.5967}\\
  \hline
  C540 \& L32 & 0.3873 & 0.3870 & 0.2083 & 0.5328\\
  \hline
  C135 \& L8 & 0.2419 & 0.2417 & 0.1114 & 0.4051\\
  \hline
\end{tabularx}
\end{center}
\end{table}

\subsection{Accuracy}
Table~\ref{tab:mAP} shows the listing for mAP, \(A_{s,l,d}\), HOTA and \(Accuracy_{norm}\) of selected sensor setups. In the sensor class camera it is noticeable that one of the medium resolutions provides comparatively better results than the highest resolution. Nevertheless, the results are substantially worse compared to the sensor class lidar. This is due to the underlying principle of the SMOKE network. SMOKE performs a depth estimation in a two-dimensional monocular image plane and  is unable to precisely determine the object's position in the direction along the optical axis of the camera, also known as the depth direction. For instance, in the case of a vehicle moving laterally, an error of as little as 1.5 meters in the depth direction can lead to a complete failure of the detection process and a mismatch with the ground truth. Further, the camera extrinsic matrix of the dataset is found to have a significant error, resulting in an inability to accurately project even correct detections and ground truth onto the correct image space. As a result, the combined setups also have slightly worse values than the lidar-only setups.

\subsection{Latency}\label{IVB}
Figure~\ref{fig:latencies} shows the results for \(t_{detection}\), \(t_{tracking}\), and \(Latency\), respectively, split by machine. The correlation of \(t_{detection}\) and the performance capabilities of the four GPU setups as well as the correlation of \(t_{tracking}\) and the performance capabilities of the three different CPU setups are illustrated. Based on the mean values of \(t_{detection}\), an improvement by a factor of 2.54 from machine no. 4 to machine no. 1 takes place on the GPU level. However, the use of two instead of one identical GPU hardly yielded any latency gains. Due to the non-parallelized execution of object tracking, single-core performance is crucial for keeping latency as low as possible. For this reason, the CPUs of machine no. 3 and 4 deliver better results than the server CPUs of machine no. 1 and 2. In the total latency, there is finally a performance leap of a factor of 1.8 between machine no. 1 and machine no. 4.
\begin{figure}[tb]
    \begin{subfigure}{0.32\columnwidth}
        \includegraphics[width=\columnwidth]{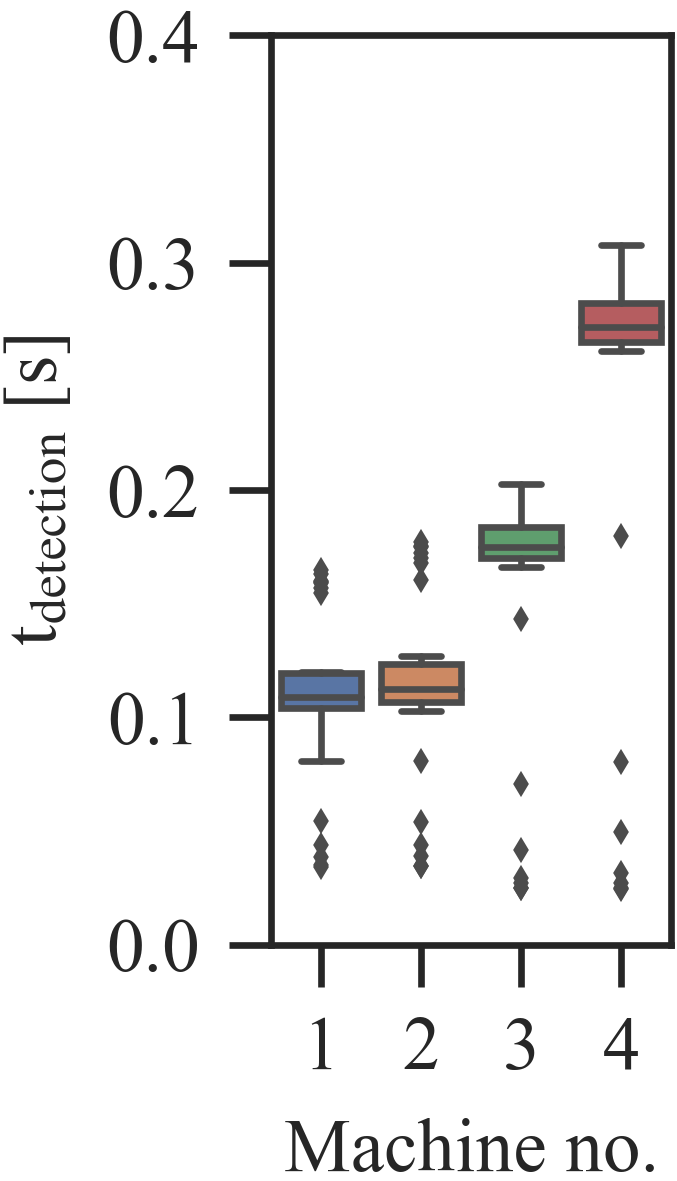}
    \end{subfigure} 
    \hfill
    \begin{subfigure}{0.32\columnwidth}
        \includegraphics[width=\columnwidth]{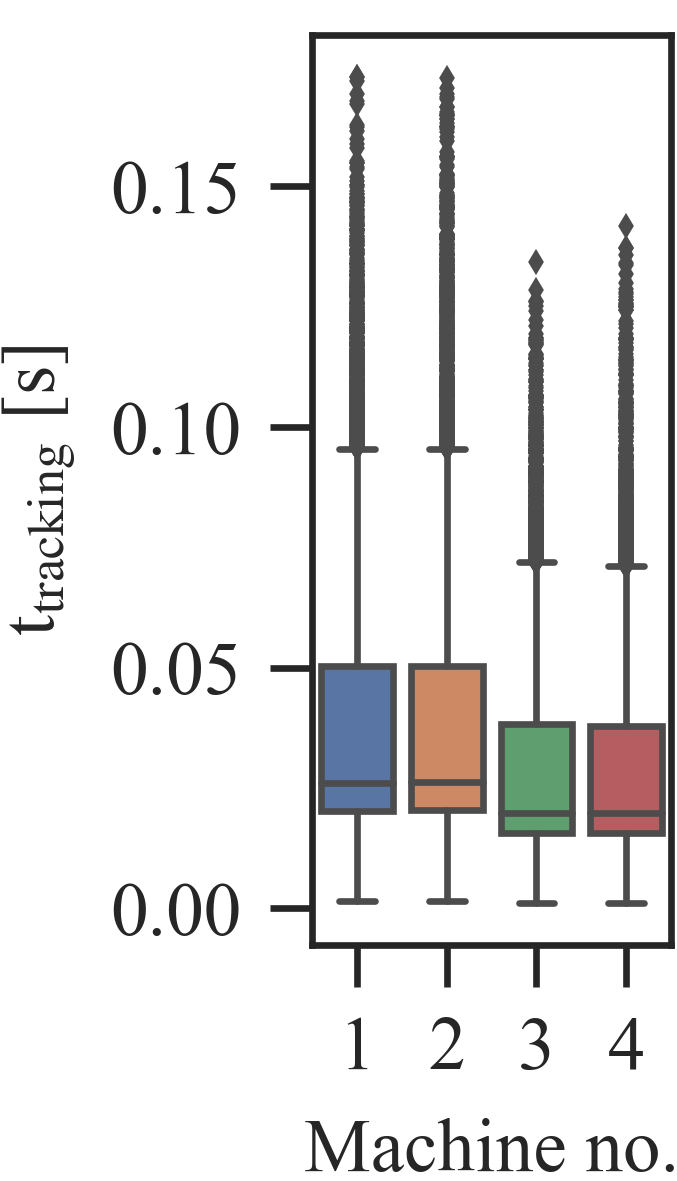}
    \end{subfigure}
    \begin{subfigure}{0.32\columnwidth}
        \includegraphics[width=\columnwidth]{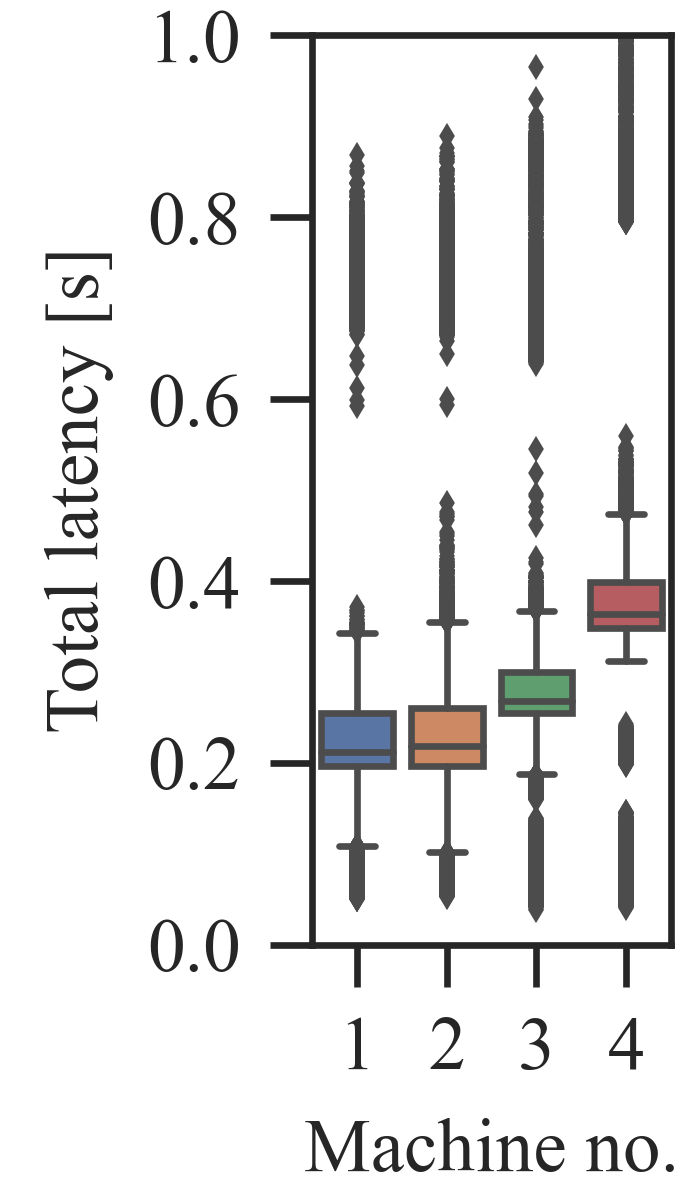}
    \end{subfigure} 
    \centering
    \caption{Representation of \(t_{detection}\), \(t_{tracking}\) and \(Latency\) of all total combinations.}
    \label{fig:latencies}
\end{figure}

\subsection{Reliability}
Figure~\ref{fig:tracking-latency-correlation} and~\ref{fig:detection-accuracy-correlation} show the distribution of \(t_{tracking}\) and \(A_d\) over the number of objects per frame. It can be seen that \(t_{tracking}\) correlates positively with the number of objects per frame. This is due to the fact that this is a serial operation, where more objects per frame contribute to a higher processing time. \(A_d\), on the other hand, correlates negatively with the number of objects per frame. The reason for this is the increasing statistical probability that not all objects are detected as the number of objects increases. A correlation between \(t_{detection}\) and the number of objects per frame could not be determined.

\begin{figure}[t]
  \begin{center}
  \includegraphics[width=\columnwidth]{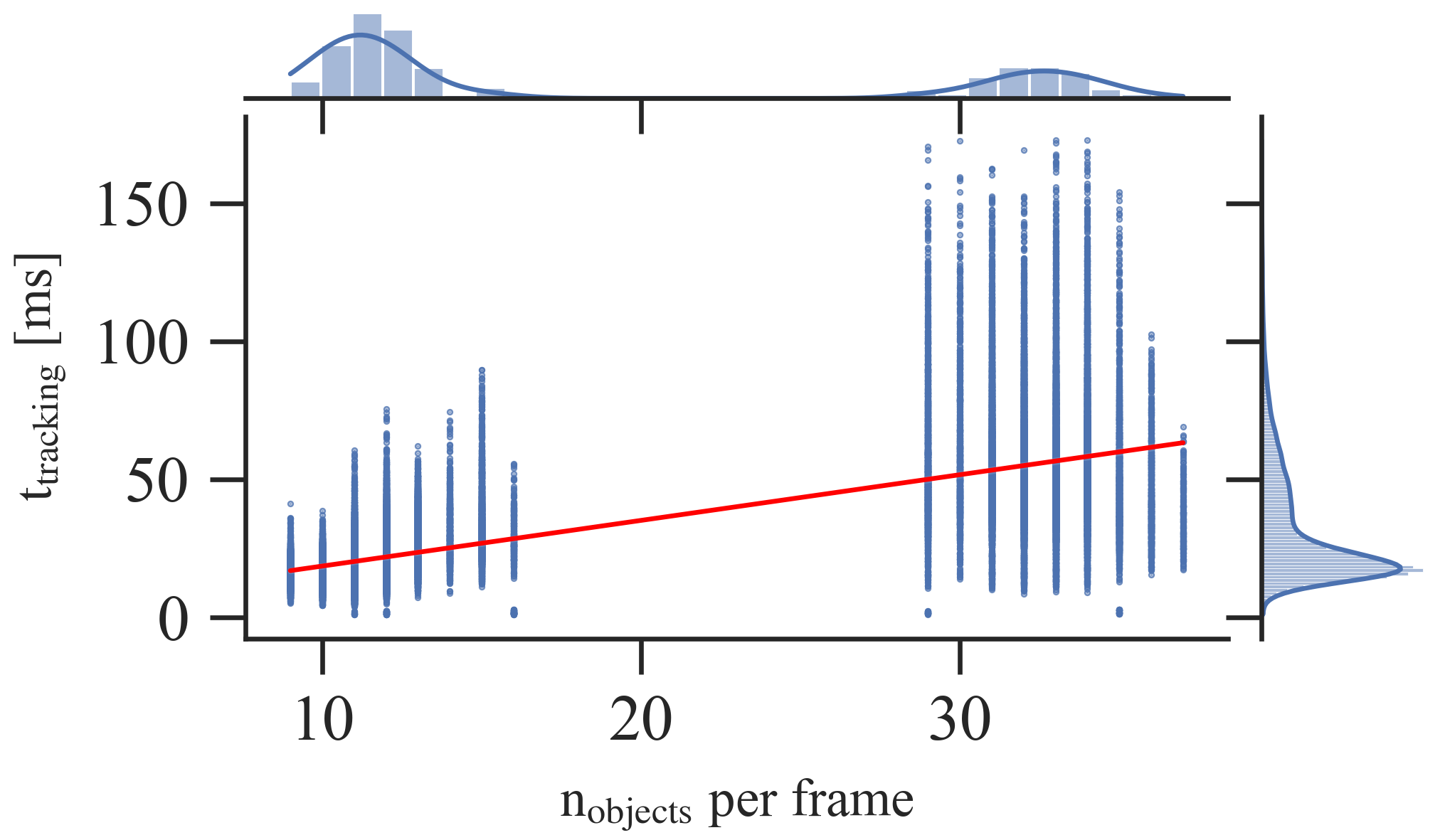}
  \caption{Distribution of \(t_{tracking}\) over the number of objects per frame.}
  \label{fig:tracking-latency-correlation}
  \end{center}
\end{figure}
\begin{figure}[t]
  \begin{center}
  \includegraphics[width=\columnwidth]{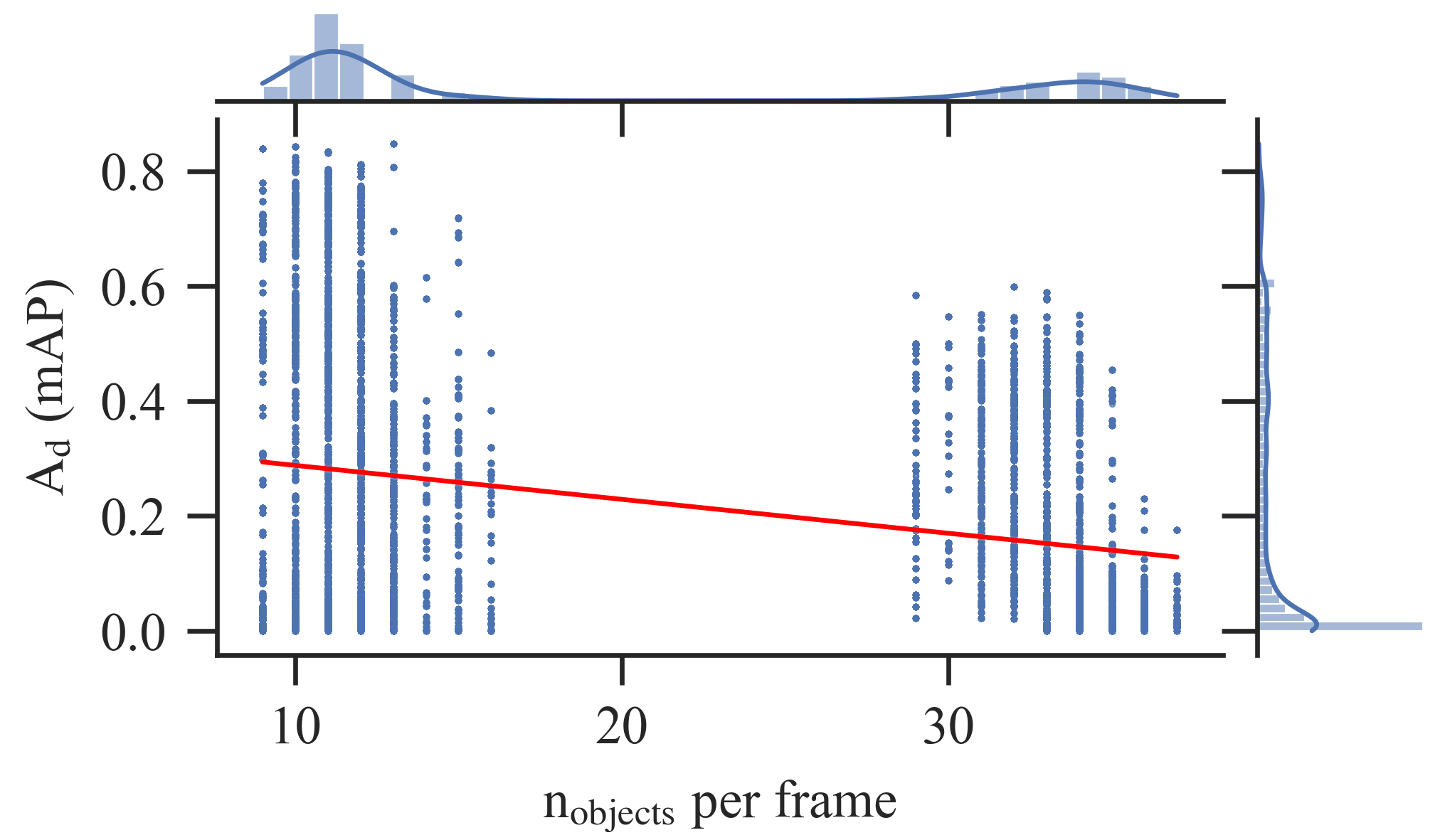}
  \caption{Distribution of \(A_d\) over the number of objects per frame.}
  \label{fig:detection-accuracy-correlation}
  \end{center}
\end{figure}

\subsection{Quality Vector}
Table~\ref{tab:ALRQ} provides an overview of \(Accuracy_{norm}\), \(Latency_{norm}\), \(Reliability_{norm}\) and \(|Q|\) for different setups. For each of these four values, the maximum and minimum is given with the corresponding setup. The highest \(Accuracy_{norm}\) is achieved with the highest lidar resolution and the lowest \(Accuracy_{norm}\) with the lowest camera resolution. For \(Latency_{norm}\), the highest result is achieved with the second lowest camera resolution on machine no. 4 and the lowest result is achieved with the combination of the highest camera and lidar resolution on machine no. 4. By definition, \(Reliability_{norm}\) ranges between 0 and 1. The value 1 is achieved with the lowest lidar resolution on machine no. 3. The value 0 is achieved with the combination of the highest camera resolution and the second highest lidar resolution on machine no. 4.

\(|Q|\) can be considered as a total score, which equally considers \(Accuracy_{norm}\), \(Latency_{norm}\) and \(Reliability_{norm}\). The best overall result is achieved with the highest lidar resolution in combination with the specifications of machine no. 1. The worst result, however, is achieved with the sensor setup consisting of the highest resolution camera and the lowest resolution lidar in combination with the specifications of machine no. 4. All 248 result combinations are additionally shown in figure~\ref{fig:ALR}.

\begin{table}[t]
\caption{Listing of maximum and minimum quality vector entries and lengths for different setups.}
\label{tab:ALRQ}
\begin{center}
\begin{tabularx}{86 mm}{ |l||X|X|X|X| }
  \hline
  \textbf{Setup} & \textbf{\(A_{norm}\)} & \textbf{\(L_{norm}\)} & \textbf{\(R_{norm}\)} & \textbf{\(|Q|\)}\\
  \hline 
  L256, machine 3 & \textbf{0.7189} & 0.9685 & 0.7184 & 1.4038\\
  \hline
  C135, machine 1 & \textbf{0.1285} & 0.9153 & 0.8672 & 1.2675\\
  \hline
  C180, machine 4 & 0.1629 & \textbf{0.9348} & 0.9628 & 1.3519\\
  \hline
  C2160 \& L256, machine 4 & 0.5974 & \textbf{0.0636} & 0.0052 & 0.6008\\
  \hline
  L8, machine 3 & 0.4392 & 0.7456 & \textbf{1.0000} & 1.3224\\
  \hline
  C2160 \& L128, machine 4 & 0.5651 & 0.0639 & \textbf{0.0000} & 0.5687\\
  \hline
  L256, machine 1 & 0.7759 & 0.9528 & 0.7184 & \textbf{1.4233}\\
  \hline
  C2160 \& L8, machine 4 & 0.0713 & 0.0308 & 0.3985 & \textbf{0.4060}\\
  \hline
\end{tabularx}
\end{center}
\end{table}

\begin{figure}[t]
  \begin{center}
  \includegraphics[width=\columnwidth]{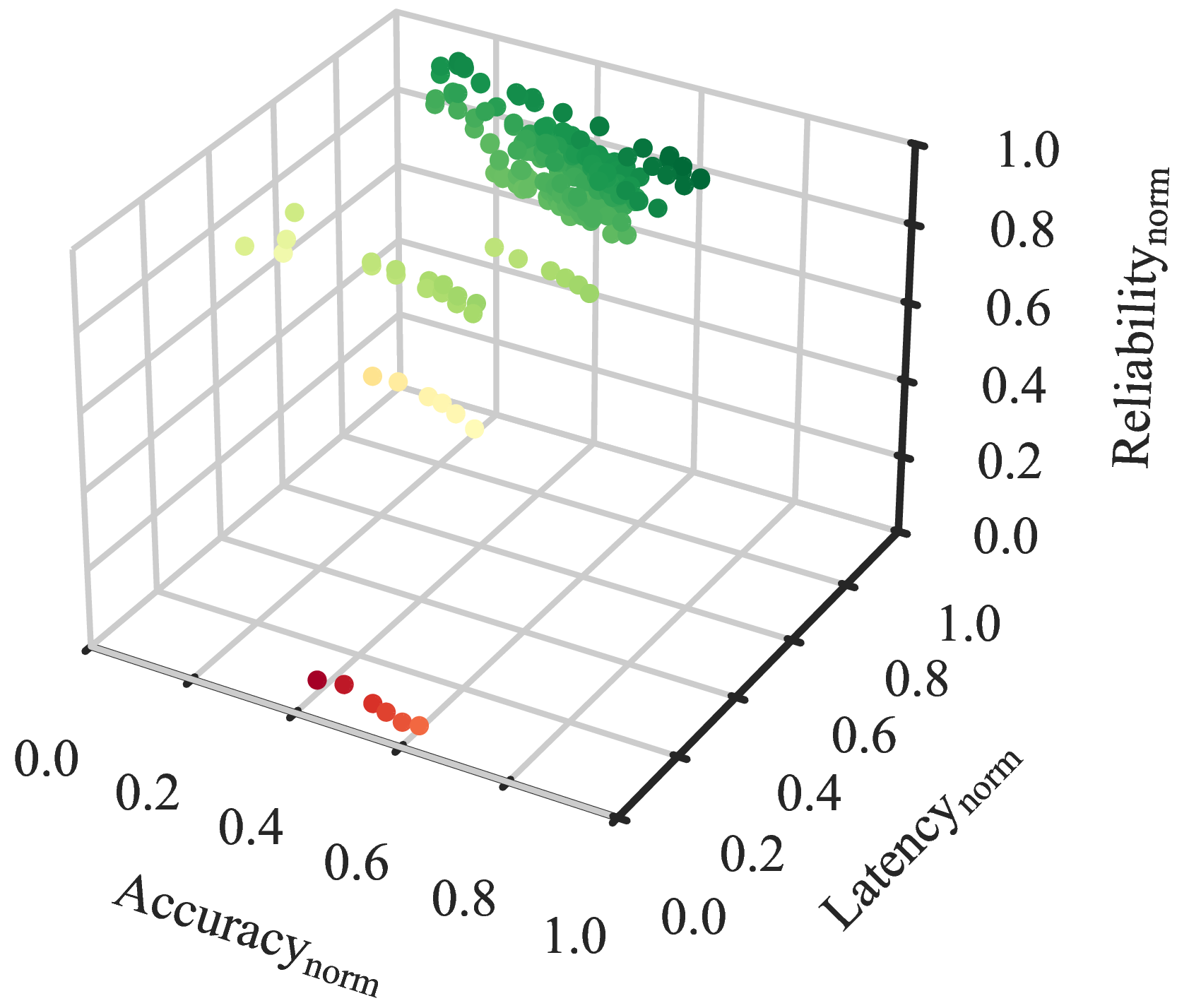}
  \caption{Display of all 248 result combinations of \(Q\). The points are colored according to their corresponding value \(|Q|\), from low (red) to high (green).}
  \label{fig:ALR}
  \end{center}
\end{figure}

In conclusion, with the sensor setups and algorithms we tested, the use of a single camera alone is insufficient for tasks requiring depth information in the infrastructure view. On average, a combined camera and lidar sensor setup is superior to a single sensor setup. The camera provides additional information about the road users in terms of dimension, orientation, and classification to the information provided by the lidar. This improves the final result, although the quality of the position estimation suffers. In terms of reliability, it was found that on lower-powered machines, lower sensor resolution is more efficient and stable, even though it comes with lower accuracy.

\section{CONCLUSIONS}
In this paper, a framework for quality assessment of intelligent roadside infrastructure sensors is presented. The framework is generically designed so that it can be applied multimodally across different sensor types. Key conclusions are made about the accuracy, latency, and reliability of investigated sensor systems consisting of one or two different sensor types. Evaluations on the DAIR-V2X dataset using four different computing environments and simulating 14 different sensor quality levels have shown that the framework provides plausible results. Previous assumptions about possible correlations between sensor and computing parameters were confirmed. A spanned quality vector space consisting of 248 infrastructure setup combinations gives so far unique insights into real ITS-S setups and their output qualities.

The analysis of further infrastructure sensor datasets allows the extension of this quality vector space. As a result, even more precise recommendations for action can be derived for current or future applications in the field of digital twins, automated driving and smart cities, regardless of the downstream function.


\addtolength{\textheight}{-20.3cm}   

\bibliographystyle{IEEEtran}
\bibliography{references}

\end{document}